\pdfoutput=1

\documentclass[11pt]{article}

\usepackage{authblk}
\usepackage{acl}
\usepackage[normalem]{ulem} 
\usepackage{float}
\usepackage{times}
\usepackage{latexsym}
\usepackage{multirow}
\usepackage{booktabs}
\usepackage[export]{adjustbox}

\usepackage{amsmath}
\usepackage{mathtools}
\usepackage{ulem}
\DeclareMathOperator*{\argmax}{arg\,max}

\usepackage{amsthm}

\theoremstyle{definition}
\newtheorem{definition}{Definition}[section]

\usepackage[T1]{fontenc}

\usepackage[utf8]{inputenc}

\usepackage{microtype}

\usepackage[textsize=scriptsize]{todonotes}

\newcommand{\method}{CBR-iKB}

\newcommand{\cbrsup}{\textsc{Cbr-kbqa}}

\title{CBR-iKB: A Case-Based Reasoning Approach for \\ Question Answering over Incomplete Knowledge Bases}

\author[*,$\dagger$]{Dung Thai}
\author[$\dagger$]{Srinivas Ravishankar}
\author[$\dagger$]{Ibrahim Abdelaziz} 
\author[*]{Mudit Chaudhary}
\author[$\dagger$]{\\Nandana Mihindukulasooriya}
\author[$\dagger$]{Tahira Naseem}
\author[*]{Rajarshi Das}
\author[$\dagger$]{\\Pavan Kapanipathi}
\author[$\dagger$]{Achille Fokoue}
\author[*]{Andrew McCallum}

\affil[*]{\small University of Massachusetts Amherst}
\affil[$\dagger$]{\small IBM Research}

\begin{document}
\maketitle
\begin{abstract}
Knowledge bases (KBs) are often incomplete and constantly changing in practice.
Yet, in many question answering applications coupled with knowledge bases, the sparse nature of KBs is often overlooked.
To this end, we propose a case-based reasoning approach, \method{}, for knowledge base question answering (KBQA) with incomplete-KB as our main focus.
Our method ensembles decisions from multiple reasoning chains with a novel nonparametric reasoning algorithm.
By design, \method{} can seamlessly adapt to changes in KBs without any task-specific training or fine-tuning.
Our method achieves 100\% accuracy on MetaQA and establishes new state-of-the-art on multiple benchmarks.
For instance, \method{} achieves an accuracy of 70\% on WebQSP under the incomplete-KB setting, outperforming the existing state-of-the-art method by 22.3\%.

\end{abstract}

\section{Introduction}
\label{sec:intro}
Knowledge base question answering (KBQA) aims to answer natural language queries using the information in Knowledge Bases (KBs).
Over the years, KBQA has attracted significant research attention~\citep{lan2021survey}, with various approaches ranging from rule-based systems~\citep{Hu2021edg}, reinforcement learning~\citep{das2018minerva}, graph query generation~\citep{shi2021transfernet} to neural semantic parsing~\citep{chen2021retrack}.


Notably, most high-performance KBQA systems~\citep{das2021cbrkbqa,ye2021rng} are tied with supervised learning, and all supporting evidence being provided in KBs.
In practice, the annotation for supervised KBQA is costly, and knowledge bases are often incomplete~\citep{min2013distant}.
Recent works~\citep{sun2018graftnet, sun2019pullnet, saxena2020embedkgqa, sun2020faithful, ren2021lego, shi2021transfernet} are designed to work on incomplete KBs, with only question-answer pairs available at training time (weakly-supervised).
While these works show promising performance gains,
the performance gap caused by incomplete-KBs still remains. 

\begin{figure}
    \centering
    \includegraphics[width=\linewidth]{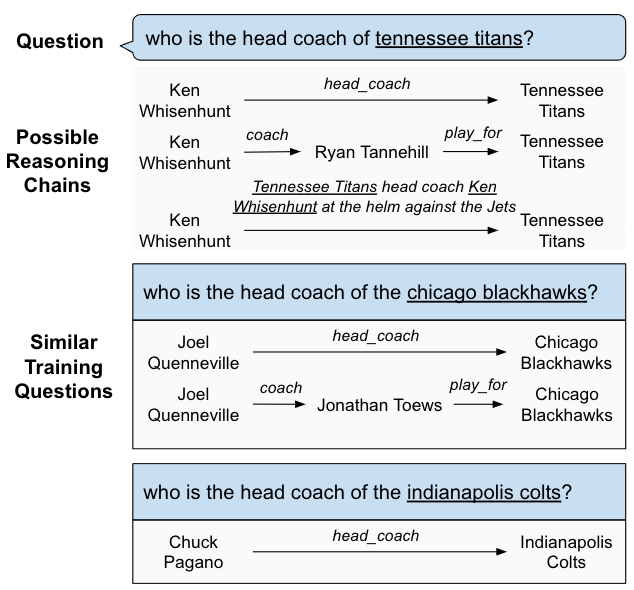}
    \caption{\textbf{An example of QA over incomplete KBs.}
    The question "who is the head coach of tennessee titans?" can be answered with several reasoning chains.
    Similar chains are available at training, but they contribute differently to models' decisions at test time.
    Holistic consideration of all correct reasoning chains is desirable for QA over incomplete KBs.
    }
    \label{fig:concept}
\end{figure}
We observe that most existing KBQA systems learn to predict the most probable reasoning chain that connects query entities and answers, which becomes problematic when KBs are incomplete.
To see why, consider the question "Who is the head coach of \emph{tennessee titans}?" and its possible reasoning chains, shown in Figure~\ref{fig:concept}.
Models trained on questions in our toy example favor the reasoning chain via relation $head\_coach$.
However, given an incomplete knowledge base, the triplet with this relation $head\_coach$ can be missing for \textit{"Ken Whisenhunt"}, causing false-positive predictions.

Inspired by the observation, we propose a novel weakly-supervised KBQA system, \textit{\method{}}, that can predict answers consistent with all possibly correct reasoning chains.
First, \method{} generates multiple reasoning chains that potentially yield answers for a newly arrived question.
Next, our method employs a majority voting scheme where each inferential chain produces voting scores for its answers.
When KB is incomplete, our method can utilize alternative reasoning chains even when (part of) some correct chains are missing.
A key design of \method{} is the integration of the case-based reasoning (CBR) paradigm with our novel nonparametric reasoning algorithm for efficiently generating reasoning chains.
CBR~\cite{Kolodner1993WhatIC,Aamodt1994CaseBasedRF} is an instance-based learning paradigm in which new problems are derived from known solutions to similar problems.
CBR-based methods are helpful for KBQA since (1) in many KBQA applications, similar questions about different entities are frequently asked, and (2) the same reasoning steps (or inferential chain) of a question can also yield correct answers to similar questions (Figure~\ref{fig:concept}).
In \method{}, we maintain a case base of questions and their inferential chains.
Given a query, \method{} uses a dense-retriever over questions' embeddings in the case base to acquire k-nearest neighbor sets of inferential chains (k-NN chains).
Due to missing triplets in the KB, some k-NN chains might be inapplicable to a new question.
Therefore, we propose a nonparametric reasoning algorithm for deriving plausible inferential chains from k-NN chains.
Our algorithm can seamlessly adapt to changes in the KB without task-specific fine-tuning.
A triplet will be automatically used in inferential chains of relevant questions whenever it is added.
This property of \method{} is desirable for applications where the KB needs to be continuously updated.

Our empirical evaluation shows that \method{} performs well on two popular KBQA benchmarks, MetaQA~\citep{zhang2018metaqa} and WebQSP~\citep{yih2016webqsp}.
\method{} achieves 100\% accuracy for MetaQA questions.
On WebQSP, as only a small fraction of questions (15\%) in the benchmark~\citep{saxena2020embedkgqa,shi2021transfernet} are not answerable with their down-sampled KB, 
we propose a more rigorous benchmark for evaluation.
In particular, we implement a triplet dropping scheme over the KB that affects half of the questions and run all methods with the new KB.
Our method significantly outperforms state-of-the-art models with incomplete-KB on this benchmark and achieves competitive performances given full-KB.

\section{Task Description}
\label{sec:task}
We consider the question-answering task where \textit{partial} background knowledge is stored in a knowledge base.
A knowledge base $K$ consists of a set of entities $E$, relations $R$, and a set of fact triplets. 
Each fact triplet is of the form $(e_s, r, e_o)$, indicating that the relation $r \in R$ exists between the subject entity $e_s \in E$ and the object entity $e_o \in E$. 
While $K$ may not cover all existing relationships between a pair of entities $(e_s, e_o)$, it is possible to infer missing relationships using a text corpus $D$.
In this work, we extend the knowledge base $K$ by adding sets of triplets of the form $(e_s, r_d, e_o)$, where $r_d$ is the relationship described in a document $d \in D$.

Now, we define relevant terminologies.
We can view the knowledge base $K$ as a graph whose nodes and edges are entities and relations, respectively.
Consider a natural language question $q$, with a linked entity $e_q$, and the target answer node $e_a$ that the KBQA system is required to find. 
Let a path $p$ from $e_q$ to $e_a$ be represented as:
\begin{align*}
p \colon e_q \xrightarrow{r_i} e_i  \xrightarrow{r_j}... \xrightarrow{r_n} e_a.
\end{align*}

\begin{definition}[Reasoning Chain]
The ordered list of entities $[e_q,...,e_a]$ and relations $[r_i,...,r_n]$ corresponding to a path $p$ is a \textit{reasoning chain}.
\end{definition}


\begin{definition}[Inferential Chain]
The ordered list of relations $[r_i,...,r_n]$ from a reasoning chain is an \textit{inferential chain}.
\end{definition}


\begin{definition}[Question Similarity]
\label{def:question_sim}
Questions $q_1$ and $q_2$ are \textit{similar} if they represent similar inferential chains but not necessarily similar reasoning chains, e.g., "Who is the head coach of Tennessee Titans?" and "Who is the head coach of the Chicago Blackhawks?".
\end{definition}


We consider the weakly-supervised setting, in which a dataset of questions $q$ and their answer sets $\{e_a\}$ is provided, but the inferential chains are not. 
We limit our setting to questions with reasoning patterns seen at training time and leave questions with novel reasoning patterns at test time for future work. 
Our task is to estimate semantically correct reasoning chains and predict the inferential chain applied to similar questions at test time. 

\begin{figure*}
    \centering
    \includegraphics[width=\linewidth]{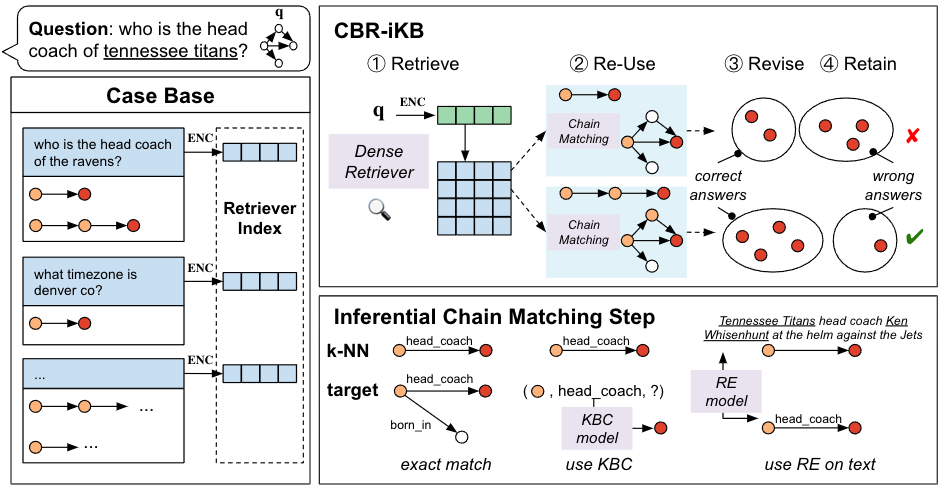}
    \caption{\textbf{Illustration of the \method{} approach.}
    The case base (left) keeps all training samples in the form of their case representations and inferential chains (solutions).
    Given a question, we first retrieve similar cases from the case base to infer all inferential chains.
    \method{} then reuses these chains (via chain matching) to produce possible answers.
    These answers are further corrected and refined in the revise and retain steps to output the final solutions.}
    \label{fig:overview}
\end{figure*}

\begin{figure}[H]
    \vspace{-5mm}
    \centering
    \includegraphics[width=0.9\linewidth]{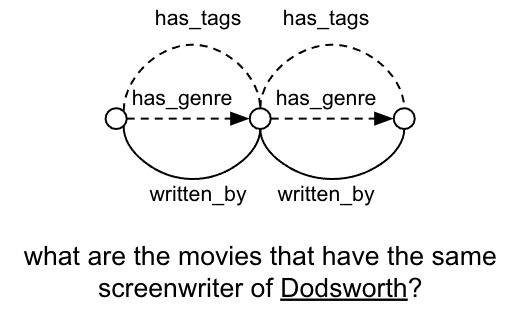}
    \vspace{-0.7em}
    \caption{A question with spurious chains (dotted arrows) and correct reasoning chain (solid arrows).}
    \label{fig:spurious}
\end{figure}

\paragraph{Inferential Chain Prediction}
\label{subsec:chain_prediction}
Given a question $q$ and its answer set $\{e_a\}$, it is straightforward to produce a set of paths ${p_i}$ between $q$ and each $e_a$.
However, not all paths are correct reasoning chains, i.e., spurious reasoning chains (Figure~\ref{fig:spurious}).
A reasoning chain is correct if its semantic behaviors are consistent with understanding the question's requirements.
Putting aside this semantically consistent property, which is hard to quantify, we observe some interesting statistical properties of the correct reasoning chains.
First, the set of correct reasoning chains usually yields the same set of inferential chains across different answers $e_a$.
Secondly, they do not introduce false-positive answers, as the spurious reasoning chains might do.
Finally, the correct reasoning chains of similar questions should also resolve to the same set of inferential chains.
We refer to this final property as the globally consistent property of the correct reasoning chains. 
We later show how to utilize these three properties to estimate the correct reasoning chains, and from there, derive the inferential chain and apply them to test questions.

\section{Proposed Method}
\label{sec:method}
Case-Based Reasoning (CBR) is an instance-based method, introduced in~\cite{schank1983cbr} and recently adapted for supervised KBQA in~\citep{das2021cbrkbqa}.
In a CBR system, training samples (or \textit{cases}) are kept in a case base. 
When a new question (or \textit{target case}) arrives, the CBR system searches the case base for similar questions (the \textit{k-nearest neighbor cases}) and their inferential chains (or \textit{solutions}).
It then reuses retrieved solutions to predict inferential chains, executes them by traversing the knowledge base, and yields desired answers.
However, the retrieved solutions are not guaranteed to be  correct and globally consistent.
Therefore, the CBR system follows up with a revise step and a retain step that refines solutions in the case base. 
In our work, the revise step computes a ranking over the solutions.
Based on this ranking, our retain step discards solutions and cases that are likely spurious. Shown in Figure~\ref{fig:overview} is an illustration of \method{}.
\subsection{The Case Base}
\label{subsec:casebase}
A CBR system operates on a case base of previously seen samples and their solutions.
This section formally defines our case base and describes how we construct it from the training dataset.

A case base $\mathcal{C}$ is a set of cases, where each case $c$ is a pair of (1) case representation $\mathbf{x}$, and (2) set of inferential chains $\mathcal{P}$. 
Given the knowledge base $K$ and a pre-trained language model $\texttt{LM}$, we can formally define $c$ as follows,
\begin{align*}
    c \coloneqq (\mathbf{x}, \mathcal{P}) \coloneqq \big(\texttt{LM}(q_{\texttt{<MASK>}}), \{\mathbf{p} \mid K, \mathcal{E}_q, \mathcal{E}_a \}\big)
\end{align*}
where $q, \mathbf{p}, \mathcal{E}_q, \mathcal{E}_a$ are the question,  the corresponding inferential chain, the answer set, and the set of extracted query entities from $q$. 
By the similarity definition~\ref{def:question_sim}, the case representation $\mathbf{x}$ should be agnostic to entities mentioned in $q$.
Therefore, we replace all tokens of entity mentions in $q$ with a special $\texttt{<MASK>}$ token from the language model.

Now, we describe how we use each question-answer sample $(q, \mathcal{E}_a)$ from the training dataset to derive a case in the case base. 
First, we use pre-trained language model $\texttt{LM}$ to encode the masked question $q_{\texttt{<MASK>}}$ and produce a case representation $\mathbf{x}$, similar to~\citep{das2021cbrkbqa}.
Next, a set of inferential chains $\mathcal{P}$ is derived from the question $q$ and the answer set $\mathcal{E}_a$.
From the question $q$, a set of query entities $\mathcal{E}_q$ is extracted, forming a set of source nodes of reasoning chains over the knowledge base $K$.
In practice, this step is accomplished by detecting the entity mentioned in $q$ and performing entity linking to the knowledge base $K$.
Since our focus is on the reasoning step, we follow the same experimental setup as~\citep{saxena2020embedkgqa,shi2021transfernet} and assume that $\mathcal{E}_q$ is given.
For each pair of query entity $e_q \in \mathcal{E}_q$ and answer entity $e_a \in \mathcal{E}_a$, we find all shortest paths between them in $K$.
Then $\mathcal{P}$ is the set of all inferential chains, each corresponding to one such path.


\subsection{Retrieving Similar Cases}
\label{subsec:retrieve}
Given a new target question $q_{\texttt{tgt}}$, the first step of our proposed CBR system is retrieving similar cases $c_{\texttt{knn}}$ from the case base $\mathcal{C}$.
To do so, we employ the dense-retriever $\texttt{FAISS}$~\cite{faiss} and populate its index with vectors of case representations in $\mathcal{C}$. {\color{red}}
We form the query $\mathbf{x}_{\texttt{tgt}}$ for the dense-retriever by encoding the target question using the same procedure and pretrained language model as we did for questions in the case base.
The dense-retriever provides a similarity ranking between the target question embedding $\textbf{x}_{\texttt{tgt}}$ and all cases in $\mathcal{C}$ based on the cosine-similarity of their embeddings.
We gather the k-nearest neighbors (k-NN) from this ranking\footnote{If there are cases with the same score then all of them will be included.}, and for each such case $c_{\texttt{knn}}$, we collect its corresponding set of inferential chains or inferential set $\mathcal{P}$ in short.
At the end of the CBR retrieve step, we obtain a collection $\{\mathcal{P}_i\}_{i=1}^k$ of the inferential sets of k-NN cases.

\subsection{Reusing Inferential Chains}
\label{subsec:reuse}
The CBR hypothesis~\citep{hllermeier2007cbrtheory} states that similar problems should have similar solutions.
In our scenario, correct inferential chains of the target question should be similar to retrieved inferential chains.
If the KB is ideal and complete, traversing the knowledge base using the same steps in retrieved inferential chains would yield desired answers.
However, the KB is often sparse in practice, resulting in different inferential chains for the same semantic behavior (see Figure~\ref{fig:concept}).
Hence, we propose an algorithm for reusing retrieved inferential chains robust to the sparsity or incompleteness of knowledge bases.

We propose a majority voting scheme where each k-NN case casts a voting score for each candidate answering node, based on its set of inferential chains.
Voting scores are aggregated across cases, and candidate nodes with the highest scores are returned as predicted answers.
Intuitively, \method{} scans all possible k-NN solutions, applies them to solve the target question, and picks answers that have high scores and appear frequently enough (the most reliable answers).

Next, we describe how \method{} computes voting scores from its inferential set $\mathcal{P}$.
Consider the target question $q_{\texttt{tgt}}$, we obtain its set of query entities $\mathcal{E}_0$, and its candidate sub-KB $K_{\texttt{tgt}}$ similar to~\citep{saxena2020embedkgqa,shi2021transfernet}. 
For each inferential chain $\mathbf{p}_{\texttt{knn}} \in \mathcal{P}$, we propose a beam search procedure that softly-following $\mathbf{p}_{\texttt{knn}}$’s relation edges on $K_{\texttt{tgt}}$. Specifically, starting from $e_0 \in \mathcal{E}_0$ and $r_0 \in \mathbf{p}_{\texttt{knn}}$, the beam search step finds a \textit{plausible} relation edge $\hat{r_0} \in K_{\texttt{tgt}}$ that matches $r_0$ and follows $\hat{r_0}$ to reach some entity nodes $e_1 \in \mathcal{E}_1$.
This beam search step is repeated for the rest of relation edge $r_i \in \mathbf{p}_{\texttt{knn}}$ in their corresponding order.
For each beam search step, a score of how likely the \textit{plausible} relation $\hat{r_i}$ holds in $K_{\texttt{tgt}}$ is also computed.
At the end of the beam search procedure, all entities in $\mathcal{E}_n$ are assigned the beam search score as their voted scores.

We employ several methods to find the \textit{plausible} relation $\hat{r_i} \in K_{\texttt{tgt}}$, depends on $r_i \in \mathbf{p}_{\texttt{knn}}$ and the target knowledge graph $K_{\texttt{tgt}}$.
By our definition of the knowledge base (in Section~\ref{sec:task}), $r_i$ can be a symbolic relation predefined by the KB or a free-form relation indicated by a short-text document.
If $r_i$ is a symbolic relation, or formally $r_i \in R$, then \method{} forms a structure query $(e_i, r_i, ?)$ \textit{over the full knowledge base} $K$.
To execute this query, we use both exact matching of $r_i$ and a pre-trained knowledge base completion model~\citep{trouillon2016complex}, notated $\texttt{KBC}$.
Here we note that this query is executed over the full KB instead of the target sub-KB $K_{\texttt{tgt}}$, allowing \method{} to consider all possible entities in the KB.

In addition, we utilize the set of free-form relations $d \in D$ that stem from $e_i$, checking whether they serve as evidence for how likely $r_i$ holds between $e_i$ and other entities mentioned in $d$.  
For this purpose, we employ an off-the-shelve relation extraction model $\texttt{RE}$~\citep{han2019opennre} specifically chosen for each benchmark (see details in Section~\ref{subsec:implementation}).
Typically, a relation extraction model predicts a relation label for an entity pair mentioned in the given text, and the set of relation labels might not be aligned with $R$.
To avoid this relation set mismatch, we suggest using a fixed proxy-text $d_{r_i}$ for all relation $r_i \in R$.
A symbolic relation $r_i \in R$ is said to be supported by the document $d \in D$ if the relation extraction model $\texttt{RE}$ predicts to the same relation given $d_{r_i}$and given $d$.
In summary, for a symbolic relation $r_i \in \mathbf{p}_{\texttt{knn}}$ and $r_i \in R$, $r_i$ plausibly holds for $e_i$ and some entities $e_{i+1} \in \mathcal{E}_{i+1}$with some score $s_i$ defined as follows,
\begin{equation}
  \resizebox{0.89\hsize}{!}{
  $
    s_i \coloneqq 
\begin{cases}
1.0 & \text{if } (e_i, r_i, e_{i+1}) \in K\\
\texttt{KBC}(e_i, r_i, e_{i+1}) &\\
\text{Pr}\big(\texttt{RE}(d_{r_i}) = \texttt{RE}(d)) & \text{if}~ (e_i, d, e_{i+1}) \in K
\end{cases}
$
}
\label{eq:si}
\end{equation}

\noindent When $r_i$ a free-form relation indicated by the document $d \in D$, we align it to a relation $r_j \in R$ using the relation extraction model. More specifically, 
\begin{align*}
    r_j \coloneqq \argmax_{r_k \in R}{ ~\text{Pr} (\texttt{RE}(d) = \texttt{RE}(d_{r_k}))}
\end{align*}
We next use $r_j$ as the plausible relation to follow from $e_i$, similar as previously described.
The scores for all entities $e_{i+1}$ resulting from following $r_j$ from $e_i$ now become,
\begin{align*}
s_i \coloneqq \text{ Pr} (\texttt{RE}(d) = \texttt{RE}(d_{r_j})) \cdot s_j
\end{align*}
where $s_j$ is computed for $r_j$ with equation (\ref{eq:si}).

\subsection{Revising and Retaining Solutions}
\label{subsec:revise}
So far, we assume that inferential chains obtained from k-nearest neighbor cases are equally correct. 
However, inferential chains are inferred from question-answers pairs and are sometimes spurious, as discussed in Section~\ref{subsec:chain_prediction}.
To alleviate the effect of spurious chains, we introduce a CBR revise step that utilizes a cross-validation set to provide a ranking over inferential chains.
Inferential chains with higher ranks are retained in the case base. 
Meanwhile, low-ranked chains with scores below a thresh-hold are considered spurious and are discarded from the case base. 

Our revise step is based on three main observations.
First, if a question has multiple answers, inferential chains should be consistent across all answers.
Here, one can see that spurious inferential chains might result in false negatives.
On the other hand, a correct inferential chain might as well introduce false negatives due to missing KB relations.
Therefore, we cannot immediately discard inferential chains with false negatives. 
Still, we can claim that the fewer false negatives are, the more reliable inferential chains are.

After extracting inferential chains from reasoning chains, we can execute inferential chains in the knowledge base.
If inferential chains are spurious, they sometimes introduce additional answers.
Ideally, this property is unique to spurious chains as correct inferential chains are bound to only correct answers.
However, some correct answers might be missing from the gold answer set due to annotation errors in practice.
These missing answers might become false positives, even for correct inferential chains.
Though false positives are not explicit indicators of spurious inferential chains, they indicate how precise inferential chains are.

Recall the CBR hypothesis that similar problems should have similar solutions. 
The two mentioned properties should hold not only for the question from which inferential chains are derived but also for similar questions.
Combining the three observations, we suggest that the F1 scores are computed for inferential chains in the case base over (1) corresponding questions derived from and (2) similar questions from a cross-validation set.
While the first set of F1 scores tells us how locally consistent inferential chains are, the second set of F1 scores lets us know how they are globally consistent with similar examples.
We rank inferential chains based on the first then the second F1 scores and retain only top inferential chains.

\section{Experiments}
\label{sec:experiments}
In this section, we compare \method{} with four other baselines on two datasets across complete and incomplete KB settings. 

\subsection{Datasets}
\label{subsec:datasets}

\begin{table}[]
    \centering
    \begin{tabular}{c|ccc}
        \hline \textbf{Dataset} & \textbf{Train} & \textbf{Dev} & \textbf{Test} \\
        \hline MetaQA 1-hop & 96,106 & 9,992 & 9,947 \\
        MetaQA 2-hop & 118,980 & 14,872 & 14,872 \\
        MetaQA 3-hop & 114,196 & 14,274 & 14,274 \\
        WebQSP & 2,848 & 250 & 1,639 \\
        \hline
    \end{tabular}
    \caption{\textbf{Dataset statistics.} We summarize the number of questions in the train, development, and test sets of MetaQA and WebQSP datasets.}
    \label{tab:dataset}
\end{table}

\textbf{MetaQA}~\cite{zhang2018metaqa} is a multi-hop QA dataset with approximately 400K questions
generated from 12 templates.
The KB contains 43K entities and 8 relations from the movie domain, with 135K triplets in total. 
Questions in MetaQA are answerable using the corpus (18K passages) provided in the original WikiMovies dataset. 



\noindent\textbf{WebQuestionsSP}~\citep{yih2016webqsp} is a multi-hop QA dataset with Freebase being its underlying KB.
It is a subset of the WebQuestions dataset~\citep{berant2013webquestions} with questions crawled from Google Suggest API.
The dataset has 4887 questions in total; each question is coupled with a topic entity, a gold inferential chain, and a set of additional constraints.
Following~\citep{saxena2020embedkgqa,shi2021transfernet}, we consider all entities within 2 hops of the entities mentioned in the question as candidate answers. 
For the text corpus, we use the Wikipedia documents set provided by GRAFT-Net~\citep{sun2018graftnet}.
GRAFT-Net retrieves the top 50 sentences relevant to query entities for each question.\\
For both datasets, the complete inferential chain required to answer each question is present in KBs.


\begin{table*}[ht!]
    \centering
    \resizebox{\textwidth}{!}{
    \begin{tabular}{c|ccc|ccc|c|c}
        \hline 
        \multirow{2}{*}{Model} & \multicolumn{3}{|c|}{ MetaQA (full) } & \multicolumn{3}{|c|}{ MetaQA (half) } & WebQSP & WebQSP \\
        \cline { 2 - 7 } & 1-hop & 2-hop & 3-hop & 1-hop & 2-hop & 3-hop & (full) & (half) \\
        \hline GRAFT-Net~\cite{sun2018open} & $97.0$ & $94.8$ & $77.7$ & $91.5$ & $69.5$ & $66.4$ & $66.4$ & $27.7$ \\
        PullNet~\cite{sun2019pullnet} & $97.0$ & $99.9$ & $91.4$ & $92.4$ & $90.4$ & $85.2$ & $68.1$ & $-$ \\
        EmbedKGQA~\cite{saxena2020embedkgqa} & $97.5$ & $98.8$ & $94.8$ & $83.9$ & $91.8$ & $70.3$ & $66.6$ & $46.7$ \\
        TransferNet~\cite{shi2021transfernet} & $97.5$ & $\boldsymbol{100}$ & $\boldsymbol{100}$ & $96.0$ & $98.5$ & $94.7$ &  $71.4$ & $47.7$ \\
        \hline \textbf{\method{} (ours)} & $\boldsymbol{100}$ & $\boldsymbol{100}$ & $\boldsymbol{100}$ & $\boldsymbol{100}$ & $\boldsymbol{100}$ & $\boldsymbol{100}$ & $\boldsymbol{78.3}$ & $\boldsymbol{70.0}$ \\
        \hline
    \end{tabular}
    }
    \caption{\textbf{Hit@1 results.}
    \method{} outperforms all other baselines over all datasets and settings.
    \method{} achieves perfectly 100\% accuracy for all settings of MetaQA, and state-of-the-art accuracies on WebQSP settings (78.3\% and 70.0\%).
    The performance gaps between \method{} and the  second-best method (TransferNet) are remarkable, especially on challenging datasets.
    For instance, these gaps are 5.3\%, 6.9\%, and 22.3\% for half-MetaQA with 3-hop, full-WebSQP, and half-WebQSP respectively, proving the significant improvement of our method.
    }
    \label{tab:main_results}
\end{table*}

\subsection{Baselines}
We compare \method{} with four baseline models: GRAFT-Net~\cite{sun2018open}, PullNet~\cite{sun2019pullnet}, EmbedKGQA~\cite{saxena2020embedkgqa}, and TransferNet~\cite{shi2021transfernet}. 
\textbf{GRAFT-Net} is a graph convolution-based approach that operates over a graph of KB triplets  and text documents.
\textbf{PullNet}~\cite{sun2019pullnet} is an improved version of GRAFT-Net with a learned CNN-based subgraph retriever.
However, its experiments are not reproducible, and we only report its numbers~\cite{shi2021transfernet}.
\textbf{EmbedKGQA} treats question embeddings as latent relation representations and jointly trains them with KB triplets.  
It is the state-of-the-art model for the WebQSP dataset with an incomplete KB.
\textbf{TransferNet} proposes a step-wise, attention-based neural network model that simultaneously traverses the knowledge graph and its alternative text form. It is state-of-the-art on both datasets with the full KB and MetaQA with the half KB. 

We report numbers on MetaQA from~\cite{shi2021transfernet} and re-run their systems on WebQSP with our proposed incomplete KB for all baselines.

\subsection{Incomplete KB Evaluation}

To simulate an incomplete-KB setting, prior works~\cite{sun2018graftnet, sun2019pullnet, saxena2020embedkgqa} randomly drop some fraction of triplets in the KB. 
However, we find that when dropping half of the triplets, much smaller (only 15\%) fractions of questions are affected.   
Thus the reported performances for the incomplete-KB setting involve many questions that are, in fact, complete.

We propose to randomly drop triplets  per question to simulate a more rigorous evaluation for incomplete KBQA, especially for a small-scale QA dataset like WebQSP that has a large-scale KB. This ensures that each question evaluates the QA systems' performance under the incomplete setting.

For each question in the dataset, we decide whether to drop its triplets  with some probability $p$.
Next, we pick a relation at random from the gold inferential chain and drop all triplets in the KB-subgraph associated with the selected relation.
We can control the fraction of questions affected by the incomplete KB by modifying $p$, which we set to $0.5$ for WebQSP. 
In addition, we continue to randomly drop triplets from the entire KB to simulate the effect of incomplete KB on knowledge base completion models. 
We intentionally keep the incomplete MetaQA baseline as-is for ease of comparison to baselines.

\subsection{Implementation Details}
\label{subsec:implementation}
\textbf{Cases Retriever.} Our cases retriever consists of a question encoder and a dense-retriever.
We use FAISS~\citep{faiss}, a standard dense-retriever for our task.
For the question encoder, we use a pre-trained DistilRoBERTa model from sentence-transformers, which has proven to provide better sentence embeddings than $\texttt{<CLS>}$ token embeddings from a language model~\cite{reimers-2019-sentence-bert}.


\noindent\textbf{Graph APIs.} Our graph traversal and handling algorithms are implemented using Graph-Tool\footnote{\url{https://graph-tool.skewed.de/}}. 
All experiments are 
run on a shared 2x Intel Xeon Silver CPU node with 1x V100 GPU.

\noindent\textbf{Knowledge Base Completion Model.} We employ the LibKGE~\citep{libkge} training, hyperparameters tuning, and evaluation pipeline.
Due to resource constraints, we only consider the ComplEx model~\citep{qin2020complex} and leave further investigations of others for future work.
We report the detailed evaluation of knowledge base completion models in Appendix~\ref{sec:appendix}.

\noindent\textbf{Relation Extraction Model.} In Section~\ref{subsec:reuse}, we propose the use of a relation extraction model for aligning k-nearest neighbor relations and relations in the target question sub-KB.
We employ the Wiki80-CNN model from the OpenNRE toolkit~\citep{han2019opennre} as our RE model.
Inputs to the relation extraction model are documents from the text corpus and the proxy-text for the relations in the KB.
For each pair of subject and object entities, the position of their mention spans is also fed into the RE model.

\subsection{Main Results}
Table~\ref{tab:main_results} presents the performance of all QA systems coupled with the full-KB and the half-KB.
The experimental results demonstrate that \method{} significantly outperforms state-of-the-art models across all sub-tasks.
On the MetaQA dataset, our method can answer all questions correctly, fully utilizing the complementary text.
On the WebQSP dataset, \method{} outperforms the state-of-the-art model (TransferNet) by 7.1\% accuracy for the full-KB setting and 22.3\% for the half-KB setting.
Here our best results are obtained on both KB and text.
Both EmbedKGQA (by design) and TransferNet (due to scalability issues) do not utilize text.
Without text, our method improved the accuracy of TransferNet by 5.3\% (full-KB) and 14\% (half-KB), compared to the accuracy reported in Table~\ref{tab:ablation_webq}.

\begin{table}[h]
    \centering
    \begin{tabular}{lccc}
        \hline 
        \multirow{2}{*}{Model} & \multicolumn{3}{c}{ Hits@1 } \\
        \cline{2-4} & 1-hop & 2-hop & 3-hop\\
        \hline \method{} & $100$ & $100$ & $100$ \\
        \hspace{3mm} 
        \textit{w/o revise} & $99.9$ & $98.7$ & $98.3$ \\
        \hspace{3mm} \textit{w/o text} & $70.9$ & $59.9$ & $86.7$ \\
        \hline
    \end{tabular}
    \caption{\textbf{Ablation study on MetaQA with half-KB.}
    The performance \method{} degrades when either the revise step or the textual knowledge is excluded. 
    In particular, without text, the accuracy reduces nearly 30\% on 1-hop MetaQA, and even worse (40\%) in the 2-hop setting.
    Disabling the revise step has less effect on \method{}, causing nearly 2\% in the worst-case scenario.
    }
    \label{tab:ablation_meta}
\end{table}

\begin{table}[]
    \centering
    \begin{tabular}{lcc}
        \hline \multirow{2}{*}{Model} & \multicolumn{2}{c}{ Hits@1 } \\
        \cline{2-3} & full & half \\
        \hline \method & $78.3$ & $70.0$ \\
        \hspace{3mm} \textit{text only} & $53.6$ & $53.6$ \\
        \hspace{3mm} \textit{KB only} & $76.7$ & $61.7$ \\
        \hline
    \end{tabular}
    \caption{\textbf{Ablation study on WebQuestionSP.}
    We observe the performance degradations of \method{}  when using only text or KB. 
    The accuracy decreases from 78.3\% and 70\% to only 53.6\% when only text is used, for full and half KBs.
    The accuracy drops are less severe if using KB, nearly 2\% and 9\% for the two settings.
    }
    \label{tab:ablation_webq}
\end{table}
\subsection{Ablation Study}
Table~\ref{tab:ablation_meta} and Table~\ref{tab:ablation_webq} show results of the ablation studies on MetaQA and WebQSP. 
\paragraph{Utilizing Text.} On MetaQA, filling missing information from text delivers the most performance gain for the incomplete-KB setting.
On WebQSP, the performance improved by 1.6\% and 8.3\% with full and half KB, showing that text becomes more valuable as the KB becomes sparser.

\paragraph{Revise and Retain Cases.}
On MetaQA, we observe that the revise and retain steps are vital to improving the last few accuracy points. Specifically, 99.9\% of train questions have spurious chains filtered out during these steps.
On the other hand, the revise and retain steps do not help for WebQSP; we conjecture that the given dev set is too small for effectively verifying chains in the train set.
\paragraph{Question Embeddings.}
We perform an analysis to understand the effects of question embeddings.
Given mention spans of topic entities, we consider both keeping them (non-masked) and replacing them with the $\texttt{<MASK>}$ token (masked).
Upon iterating through k-nearest neighbors results, we observe that masking out mention spans is more desirable.
The neighborhood of questions represented with masked mentions tends to yield similar gold inferential chains.
We present some selected questions with their masked and non-masked retrievals and show them in Table~\ref{tab:ablation_mask} in the Appendix.

\section{Related Work}
\label{sec:related_work}
Our work shares goals with other approaches to improve question answering systems over incomplete knowledge bases~\citep{sun2018graftnet,sun2019pullnet,xiong2019kar, saxena2020embedkgqa,sun2020faithful,ren2021lego,shi2021transfernet}.
They explore various methods to incorporate text and predict plausibly missing KB facts.
GRAFT-Net~\citep{sun2018graftnet} proposes an approach for extracting answers from question-specific subgraphs containing text, KB entities, and relations using graph
representation learning. 
Similarly, PullNet~\citep{sun2019pullnet} uses an iterative process to construct a question-specific
subgraph that contains information relevant to the
question from the KB and text then uses a graph CNN to extract the answer. 
Nevertheless, none of these methods uses the question similarity to find similar reasoning chains.
Knowledge-Aware Reader~\citep{xiong2019kar} proposes a subgraph reader that enhances question embeddings with KB embeddings.
TransferNet~\citep{shi2021transfernet} simultaneously traverses the KB and a relation graph constructed from linked text to predict reasoning chains.
EmbedKGQA~\citep{saxena2020embedkgqa} jointly trains question and relation embeddings with a link prediction objective.
EmQL~\citep{sun2020faithful,ren2021lego} defines KB operations and performs reasoning over the latent space of KB embeddings.
These methods require task-specific training and must be fine-tuned to adapt to new facts to the KB.
Our method follows the CBR paradigm and suggests that KBQA reasoning chains can be obtained from similar examples with a nonparametric algorithm.
Our method also has access to multiple inferential chains at the inference time.
We show that our method can explicitly utilize alternative chains when KB facts are missing. 
In this regard, our method is closely related to a concurrent work~\citep{qin2020complex}, which trains to assign high probabilities to correct reasoning paths.
\method{}, on the other hand, takes a further step and aggregates predictions from multiple chains.

Case-based reasoning has been successfully adapted for various tasks~\citep{Watson1997ApplyingCR,Li2018DeepLF}, including KBQA.
Recently, \cbrsup{}~\citep{das2021cbrkbqa} proposes to generate KB queries from label queries of similar questions.
While \cbrsup{} requires full supervision, our method needs only question-answer pairs.
\cbrsup{} also proposes a revise step to correct missing relations in predicted KB queries where they fail to execute.
However, it does not fill in missing KB facts, which are common in incomplete KBs.










\section{Conclusion}
We proposed \method{}, a nonparametric and instance-based method for question answering over knowledge bases.
\method{} utilizes the case-based reasoning paradigm with a novel nonparametric reasoning algorithm efficiently  ensemble decisions from multiple reasoning chains.
Our method performs well on multiple KBQA benchmarks~\citep{zhang2018metaqa,yih2016webqsp,saxena2020embedkgqa}, even when coupled with sparse, incomplete KBs.
\method{} consistently achieves 100\% accuracy on different settings of the MetaQA dataset.
On WebQSP, our method significantly outperforms state-of-the-art models for question answering over an incomplete knowledge base by a large accuracy gap of 22.3\%.
Furthermore, our qualitative analysis also demonstrates that \method{}'s predictions are interpretable and explainable.


\paragraph{Limitations}
\method{} currently has limited generalization ability to novel compositional questions due to the assumption that solutions to a question are previously seen for similar questions.
Enabling compositional QA for \method{} is an interesting and open problem for future work.

\newpage
\bibliography{anthology,custom}
\bibliographystyle{acl_natbib}

\newpage
\onecolumn
\appendix

\section{Appendix}
\label{sec:appendix}

\begin{table*}[ht]
\renewcommand{\arraystretch}{1.2}
\centering
\resizebox{\textwidth}{!}{%
\begin{tabular}{llll}
\hline
\multicolumn{4}{|c|}{\textbf{Query:} Who are the directors of the movies written by {[}Peter Facinelli{]}}                                                                               \\ \hline
\multicolumn{2}{|c|}{\textbf{Top Masked Retrievals}}                              & \multicolumn{2}{c|}{\textbf{Top Unmasked Retrievals}}                                     \\ \hline
\multicolumn{2}{|l|}{Which person directed the films acted by {[}Jeff Fahey{]}}   & \multicolumn{2}{l|}{Which person directed the films acted by {[}Peter Facinelli{]}}       \\ \hline
\multicolumn{2}{|l|}{Which person directed the films acted by {[}Damian Lewis{]}} & \multicolumn{2}{l|}{Who are the directors of the movies written by {[}Peter Facinelli{]}} \\ \hline
\multicolumn{4}{|c|}{\textbf{Query:} In what country is {[}Amsterdam{]}}                                                                                                               \\ \hline
\multicolumn{2}{|c|}{\textbf{Top Masked Retrievals}}                              & \multicolumn{2}{c|}{\textbf{Top Unmasked Retrievals}}                                     \\ \hline
\multicolumn{2}{|l|}{What country is {[}Vatican city{]} in}                       & \multicolumn{2}{l|}{What do people go to {[}Amsterdam{]} for}                             \\ \hline
\multicolumn{2}{|l|}{What country is the {[}Grand Bahama island{]} in}            & \multicolumn{2}{l|}{Where is {[}Amsterdam{]} ohio}                                        \\ \hline
\multicolumn{4}{|c|}{\textbf{Query:} Where is the best place to vacation in the {[}Dominican Republic{]}}                                                                              \\ \hline
\multicolumn{2}{|c|}{\textbf{Top Masked Retrievals}}                              & \multicolumn{2}{c|}{\textbf{Top Unmasked Retrievals}}                                     \\ \hline
\multicolumn{2}{|l|}{Where to go in {[}Florida{]} for vacation}                   & \multicolumn{2}{l|}{What is the {[}Dominican Republic{]} 's capital}                      \\ \hline
\multicolumn{2}{|l|}{What are the best places to go in {[}Germany{]}}               & \multicolumn{2}{l|}{What currency is best to take to {[}Dominican Republic{]}}             \\ \hline
\end{tabular}%
}
\caption{Case retrieval examples on masked and unmasked questions. Entities are enclosed in square brackets.}
\label{tab:ablation_mask}
\end{table*}

\end{document}